\documentclass{article} 
\usepackage{nips12submit_e,times}

\usepackage{bm}
\usepackage{amsmath}
\usepackage{amsfonts}
\usepackage{amssymb}
\usepackage{graphicx} 
\usepackage{array}
\usepackage{multirow}
\usepackage[numbers]{natbib}

\title{Feature grouping from spatially constrained multiplicative interaction}

\author{
Felix Bauer \\
Frankfurt Institute for Advanced Studies\\
Frankfurt, Germany\\
\texttt{fbauer@fias.uni-frankfurt.de} \\
\And
Roland Memisevic\\
University of Montreal\\
Montreal, Canada \\
\texttt{memisevr@iro.umontreal.ca} \\
}

\nipsfinalcopy 

\begin{document}
\maketitle
\begin{abstract}
We present a feature learning model that learns to encode relationships between images. 
The model is defined as a Gated Boltzmann Machine, which is constrained such that 
hidden units that are nearby in space can gate each other's connections. 
We show how frequency/orientation ``columns'' as well as topographic filter maps 
follow naturally from training the model on image pairs. 
The model also offers a simple explanation why group sparse coding and topographic 
feature learning yields features 
that tend to by grouped according to \emph{frequency, orientation} and \emph{position}
but not according to phase.
Experimental results on synthetic image transformations show that spatially 
constrained gating is an effective way to reduce the number of parameters and thereby 
to regularize a transformation-learning model.  
\end{abstract}

\section{Introduction}
\label{section:introduction}
Feature-learning methods have started to become a standard component in many computer-vision 
pipelines, because they can generate representations which are better at encoding the 
content of images than raw images themselves. Feature learning works by projecting 
local image patches onto a set of feature vectors (aka. ``filters''), and using the vector 
of filter responses as the representation of the patch. This representation gets passed on 
to further processing modules like spatial pooling and classification. 
Filters can be learned using a variety of criteria, 
including maximization of sparseness across filter responses \cite{OlshausenField_nature}, 
minimizing reconstruction error \cite{denoisingAE}, 
maximizing likelihood \cite{cd}, and many others. 
Under any of these learning criteria, Gabor features typically emerge 
when training on natural image patches. 

There has been an increasing interest recently in imposing \emph{group structure} 
on learned filters. 
For learning, filters are encouraged to come in small groups, such that all members 
of a group share certain properties. 
The motivation for this is that group structure can explain several biological 
phenomena such as the presence of complex cells \cite{HyvarinenISA}, it 
provides a simple way to model dependencies between features 
(eg., \cite{jenatonneProximal} and references therein), 
and it can make learned representations more robust to small transformations 
which is useful for recognition \cite{korayCvpr09}. 
Filter grouping is referred to also as ``structured sparse coding'' or 
``group sparse coding''. 
Feature grouping can also be used as a way to obtain 
\emph{topographic feature maps} \cite{NaturalImageStatistics}. 
To this end, features are layed out in a 2-dimensional 
grid, and groups are defined on this grid such that each filter group shares filters 
with its neighboring groups. In other words, groups overlap with each other. 
Training feature grouping and topographic models on natural image patches 
typically yields Gabor filters whose frequency, orientation and position 
is similar for all the filters within a group. 
Phase, in contrast, tends to vary randomly across the filters within a group (eg. \cite{HyvarinenHI00,korayCvpr09,NaturalImageStatistics}). 

Various approaches to performing group sparse coding and topographic feature learning 
have been proposed. Practically all of these are based on the same recipe: 
The set of filters is pre-partitioned into groups before training. Then during 
learning, a second layer computes a weighted sum over the \emph{squares} of all filter responses within a group
(eg., \cite{HyvarinenISA,jenatonneProximal,korayCvpr09}).
The motivation for using square-pooling architectures to learn group structure 
is based in part on empirical 
evidence (it seems to work across many models and learning objectives).
There have also been various attempts to explain it, based on a variety of heuristics:
One explanation, for example, follows from the fact 
that \emph{some} nonlinearity must be used before pooling features in a group, because 
in the absence of a nonlinearity, we would wind up with a standard (linear) 
feature learning model. The square is a non-linearity that is simple and canonical.  
It can also be shown that even-symmetric functions, like the square, applied to 
filter responses, show strong dependencies. 
So they seem to capture a lot of what we are missing after performing a single 
layer of feature learning \cite{NaturalImageStatistics}. 
Another motivation is that Gabor features are local Fourier components, so computing 
squares is like computing spectral energies. 
Finally, it can be shown that in the presence of a upstream square-root 
non-linearity, using squared features generalizes standard feature learning, 
since it degenerates to a standard feature-learning model when using 
group size $1$ \cite{korayCvpr09}.  
For a summary of these various heuristics, see \cite{NaturalImageStatistics} (page 215).  
Topographic grouping of frequency, orientation and position is also a well-known feature 
of mammalian brains (eg., \cite{NaturalImageStatistics}). 
Thus, another motivation for using square-pooling in feature learning in general 
has been that, by means of replicating this effect, it may yield models that are 
biologically consistent. 

In this work, we show that a natural motivation for the use of squared 
filter responses can be derived from the perspective of 
encoding \emph{relationships} between images. 
In particular, we show that the emergence of group structure and topography 
follows automatically from the computation of binocular disparity or motion, 
if we assume that neurons that are nearby in space exert multiplicative 
influences on each other. 
Our work is based on the close relationship between the well-known 
``energy models'' of motion  and binocularity \cite{adelson1985spatiotemporal,ODF} 
and the equivalent ``cross-correlation'' models \cite{FleetBinocular,multiview}.  
It may help shed light onto the close relationship between topographic organization 
and motion processing as well as binocular vision. 
That way, it may also help shed light onto the phenomenon that topographic filter maps 
do not seem to be present in rodents \cite{stevens2011universal}. 

\section{Factored Gated Boltzmann Machine}
\label{section:fgbm}
While feature learning has been applied predominantly to single, static images in the past, 
there has been an increasing interest recently in learning features to encode relationships 
between multiple images, for example, to encode motion (eg., \cite{LeZouYoungNg,tayloreccv10}). 
We focus in this work on the Gated Boltzmann Machine (GBM) \cite{imtrans,tayloreccv10}
which models the relationship between two binary images $\bm{x}$ and $\bm{y}$ 
using the three-way energy function
\begin{equation}
E(\bm{x}, \bm{y}, \bm{h}) = \sum_{ijk} w_{ijk} x_i y_j h_k
\label{equation:gbm}
\end{equation}
The energy gets exponentiated and normalized to define the probability over image 
pairs (we drop any bias-terms here to avoid clutter):
\begin{eqnarray}
p(\bm{x}, \bm{y})= \frac{1}{Z} \sum_{\bm{h}} \exp\big( E(\bm{x}, \bm{y}, \bm{h}) \big), \;
Z= \sum_{\bm{x}, \bm{y}, \bm{h}} \exp\big(E(\bm{x}, \bm{y}, \bm{h})\big)
\end{eqnarray}
By adding appropriate penalty terms to the log-likelihood, one can extend the model to learn  
real-valued images \cite{imtrans}. 

\begin{figure}
    \begin{center}
    \begin{tabular}{ccc}
        \includegraphics[angle=0, width=4.0cm]{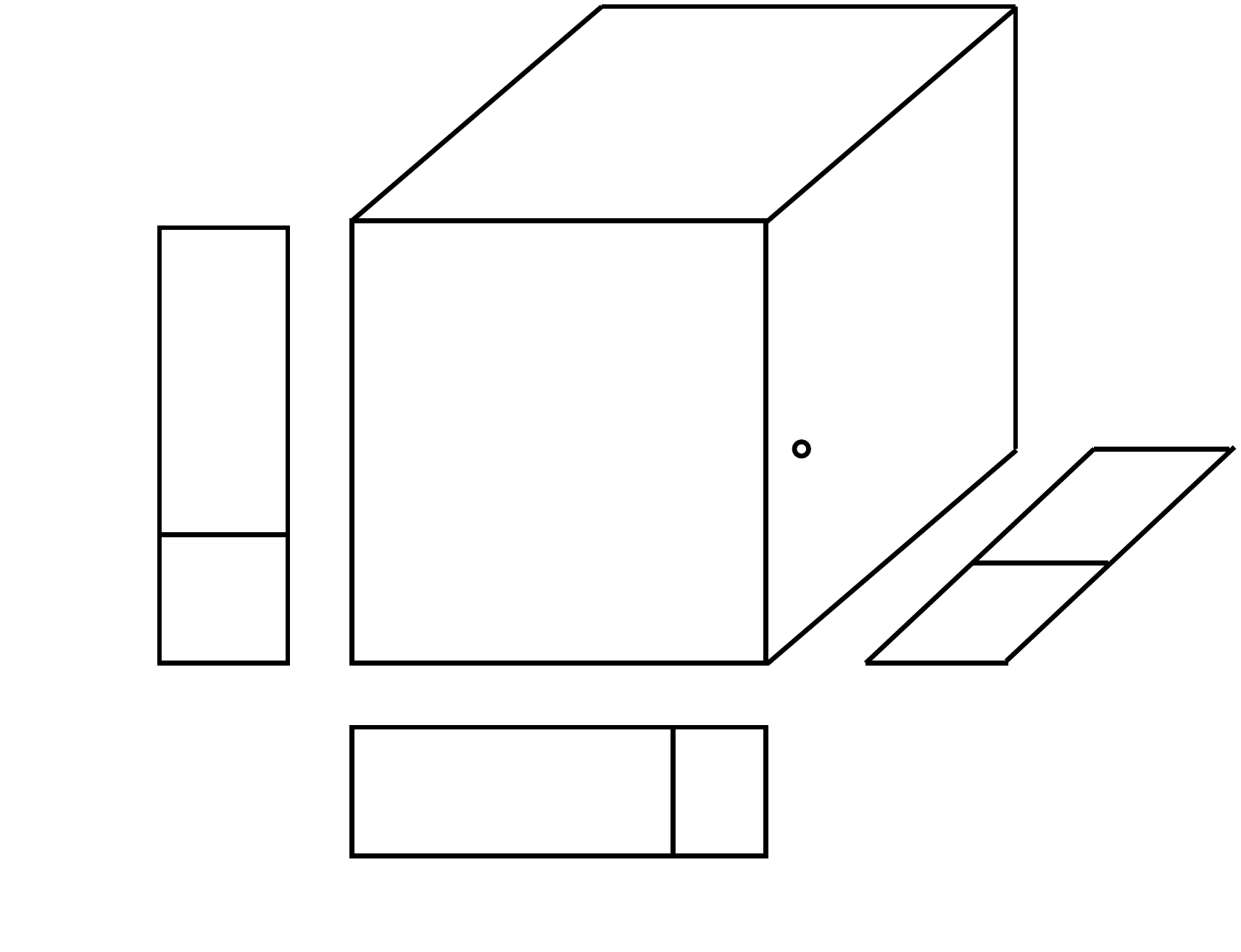}&
	\includegraphics[angle=0, width=4.0cm]{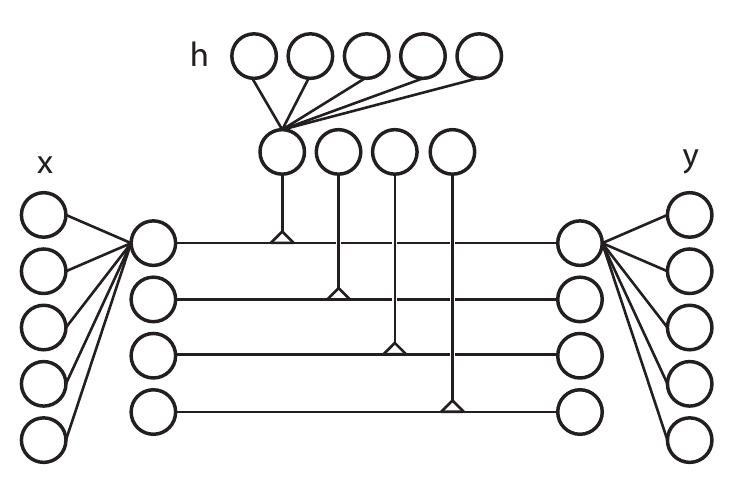} &
	\includegraphics[angle=0, width=4.0cm]{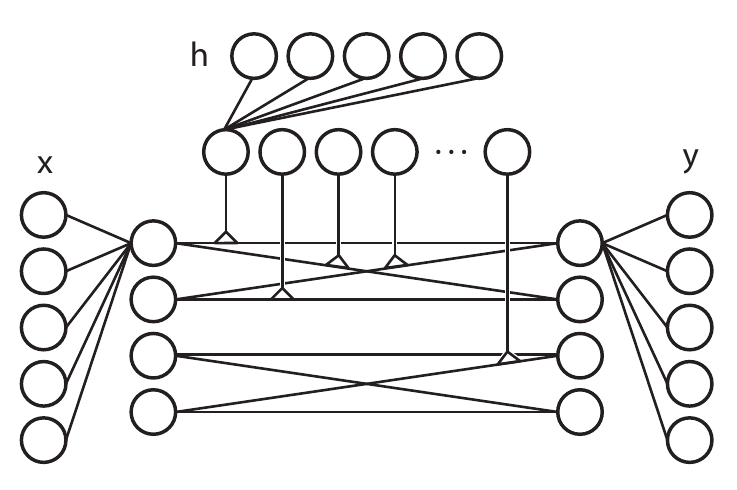}
    \end{tabular}
    \end{center}
\caption{Factorizing the parameter tensor of a Gated Boltzmann Machine ({\bf left}) is equivalent to  
gating filter responses instead of raw pixels ({\bf middle}).
In a group-gating model ({\bf right}), filters are grouped, and products of all filters within a group are 
passed on to the mapping units. In the example, there are two groups, resulting in four unique products per group.}
\label{figure:blocktensor}
\end{figure}

Since there is a product involving every triplet of an input pixel, an output pixel and a mapping unit, the number of parameters is roughly cubic in the number of pixels. 
To reduce that number, \cite{factoredGBM} suggested factorizing the three-way parameter tensor
$W$ with entries $w_{ijk}$ in Equation~\ref{equation:gbm} into a three-way inner product:
\begin{equation} 
w_{ijk} = \sum_f w^x_{if} w^y_{jf} w^h_{kf}
\label{equation:diagonalfactorization}
\end{equation} 
Here, $f$ is a latent dimension that has to be chosen by hand or by cross-validation. 
This form of tensor factorization is also known as PARAFAC or ``canonical decomposition'' 
in the literature, and it can be viewed as a three-way generalization of the SVD \cite{CarrollChang}.
It is illustrated in Figure~\ref{figure:blocktensor} (left).
The factorization makes use of a diagonal core tensor, that contains ones 
along its diagonal and that is zero elsewhere. 
Plugging in the factorized representation for $W$ and using the distributive law yields the factorized 
energy function:
\begin{equation} 
E = \sum_f \big( \sum_i w^x_{if} x_i \big) \big( \sum_j w^y_{jf} y_j \big) \big( \sum_k w^h_{kf} h_k \big)
\label{equation:energydiagonalfactorization1}
\end{equation} 

Inferring the transformation $\bm{h}$, given two images, $\bm{x}$ and $\bm{y}$, 
is efficient, because the hidden variables are independent, given the data \cite{factoredGBM}. 
More specifically we have 
$p(\bm{h}|\bm{x}, \bm{y})=\prod_k p(h_k|\bm{x}, \bm{y})$ with 
\begin{equation} 
p(h_k|\bm{x}, \bm{y}) = \sigma\big(\sum_f w^h_{kf} \big( \sum_i w^x_{if} x_i \big) \big( \sum_j w^y_{jf} y_j \big) \big),
\label{equation:inference}
\end{equation} 
where $\sigma$ is the logistic sigmoid $\sigma(z)=\frac{1}{1+\exp(-z)}$. 
Thus, to perform inference, images are projected onto $F$ basis functions (``filters'') and 
those basis functions that are in correspondence (i.e., that have the same index $f$) are multiplied. 
Finally, each hidden unit, $h_k$, receives a weighted sum over all products as input. 
An illustration is shown in Figure~\ref{figure:blocktensor} (middle, right).

It is important to note that projections onto filters typically do \emph{not} 
serve to reduce the dimensionality of inputs. Rather, it is the restricted connectivity in 
the projected space that leads to the reduction in the number of parameters. 
In fact, it is not uncommon to use a number of factors that is larger than the 
dimensionality of the input data (for example, \cite{factoredGBM, mcrbm}). 

To train the model, one can use contrastive divergence \cite{factoredGBM}, score-matching 
or a variety of other approximations. Recently, \cite{ICML2011Swersky_622, higherordergradientbased}
showed that one may equivalently add a decoder network, effectively turning the model into a
``gated'' version of a de-noising auto-encoder \cite{denoisingAE} and train it using 
back-prop. Inference is then still the same as in a Boltzmann machine.  
We use this approach in most of our experiments, after verifying that the 
performance is similar to contrastive divergence. 
More specifically, for a training image pair $(\bm{x}, \bm{y})$, let 
$\bm{p}_h(\bm{x},\bm{y})$ denote the vector of inferred hidden probabilities 
(Eq.~\ref{equation:inference}) and let $W_x, W_y, W_h$ denote the matrices 
containing the input filters, output filters, and hidden filters, 
respectively (stacked column-wise). Training now amounts to minimizing the 
average reconstruction error $\big(\bm{y}-\hat{\bm{y}}(\bm{x})\big)^2 + \big(\bm{x}-\hat{\bm{x}}(\bm{y})\big)^2$, with 
\begin{equation}
\hat{\bm{y}}(\bm{x}) = W_y \left( \Big(W_h^\mathrm{T}\bm{p}_h(\bm{x},\bm{y})\Big) * \Big(W_x^\mathrm{T}\bm{x}\Big) \right),
\end{equation}
where $*$ denotes element-wise multiplication. $\hat{\bm{x}}(\bm{y})$ is defined analogously, with all occurrences of x and y exchanged.
One may add noise to the data during training (but reconstruct the original, not noisy, 
data to compute the cost) \cite{denoisingAE}. We observed that this helps localize 
filters on natural video, but on synthetic data (shifts and rotations) it is possible to train
without noise.

\subsection{Phase usage in modeling transformations}
\label{section:phaseanalysis}
\begin{figure}[t]
\begin{center}
\begin{tabular}{cc}
  \begin{tabular}{cc}
    \includegraphics[angle=0, width=2.6cm]{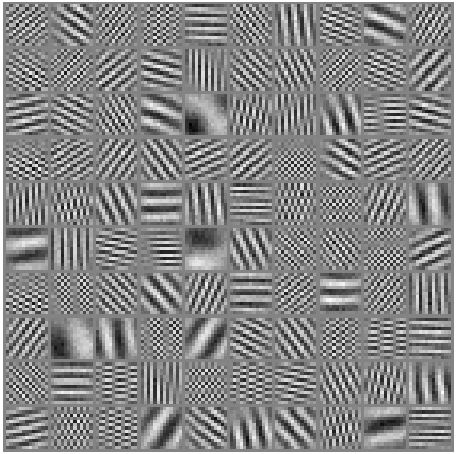}&
    \includegraphics[angle=0, width=2.6cm]{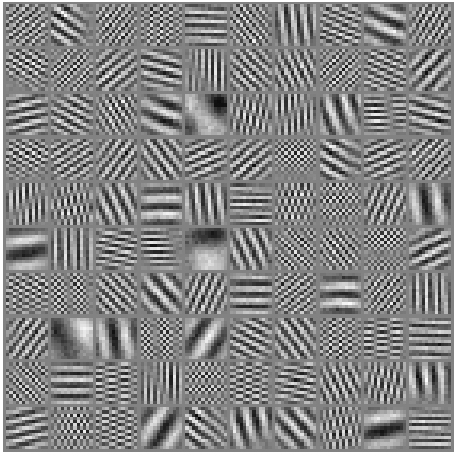} \\
    \multicolumn{2}{c}{\includegraphics[angle=0, width=5.7cm]{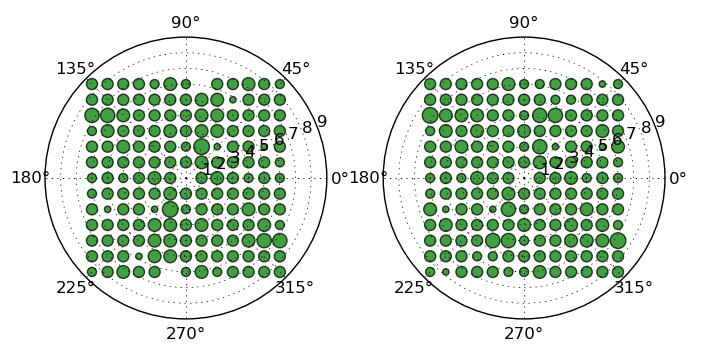}} 
  \end{tabular}
  \begin{tabular}{c}
    \hspace{-5pt}\includegraphics[angle=0, width=6.6cm]{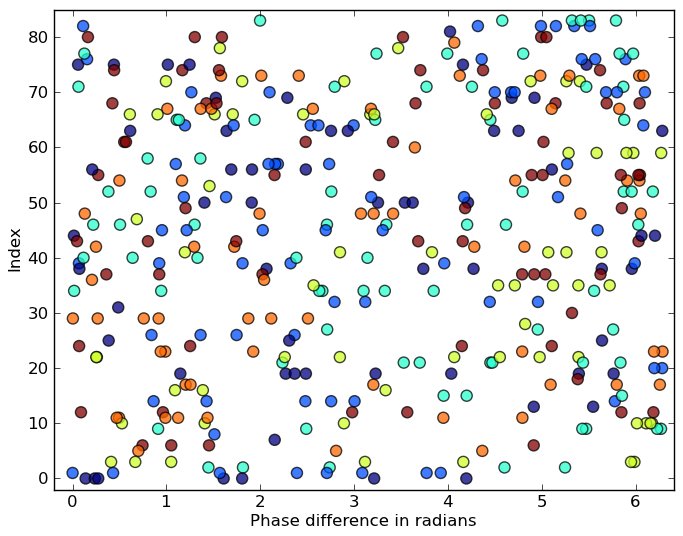}
  \end{tabular}  
\end{tabular}
\end{center}
\caption{{\bf Left, top:} Filter pairs learned from translated random-dot images. 
(Input filters on the left, output filters on the right).
{\bf Left, bottom:} Corresponding histograms showing the number of filters per frequency/orientation bin.
The size of each blob is proportional to the number of filters in each bin. 
{\bf Right:} Phase differences between learned filters. Each row corresponds to one frequency/orientation bin.
Colors are used to simplify visibility.}
\label{figure:fouriers}
\end{figure}

Figure~\ref{figure:fouriers} (left, top) shows filter-pairs that were learned from translated random-dot images using 
a Factored Gated Boltzmann Machine (for details on training, see Section~\ref{section:experiments} below). 
Training on translations turns filters into Fourier components, as was initially observed by \cite{factoredGBM}.
The left-bottom plot shows histograms over the occurrence of frequencies and orientations for input- and output-image
filters, respectively. These were generated by first performing a 2D-DFT on each filter and then picking the frequency 
and orientation of the strongest component for the filter.
The histograms show that the learned filters evenly cover the space of frequencies and orientations. 
This is to be expected, as all frequencies and orientations contribute equally to the set of random translations 
(e.g., \cite{GonzalezWoods}). 

It is also well-known, however, that multiple different translations will affect each frequency and orientation 
differently. More specifically, any given translation
induces a set of phase-shifts for every component of a given frequency and orientation. 
The phase-shift depends linearly on frequency \cite{GonzalezWoods}. 
Likewise, two different image translations of the same orientation will induce two different phase-shifts 
for each frequency/orientation. 
In order to represent translations, it is necessary to specify the phase at every frequency 
and orientation. This shows that mapping units in a GBM must have access to multiple different
phase-shifts at every frequency and orientation to be able to represent translations.  
Also, there is no need to multiply filter responses of different frequencies and/or orientations, 
only of different phases.

As a repercussion, for each input filter in Figure~\ref{figure:fouriers} (left, top) there need to be
multiple phase-shifted copies present in the output-filter bank, so that filters with varying phase
differences can be matched (which means the filters' responses will be multiplied with each other).
Likewise for each output filter. 
When the number of factors is small, the model has to find a compromise, for example, by 
connecting hidden variables, $h_k$, to multiple phase-shifted versions of filters 
with the correct phase-shift but with only a \emph{similar}, but not the same frequency and orientation.
Figure~\ref{figure:fouriers} (right) depicts the occurrences of phase differences between input- 
and output-image filters. 
It shows that the model learned to use a variety of phase differences at each frequency/orientation 
to represent the training transformations. 
In the model, each phase difference corresponds to exactly one filter pair.  

The analysis generalizes to other transformations, such as rotations or natural videos (e.g., \cite{multiview}).
In particular, it also generalizes to Gabor features which are localized Fourier features that 
emerge when training the GBM on natural video (e.g., \cite{tayloreccv10}).

\section{Group gating}
\label{groupgating}
The analysis in the previous section strongly suggests using a richer connectivity that supports the 
\emph{re-use} of filters. 
In this case the model can learn to match any filter $\bm{w}^x_{f}$ with multiple phase-shifted 
copies $\bm{w}^y_{f}$ of itself rather than with a single one. 
All phase differences in Figure~\ref{figure:fouriers} (right) can then in principle be 
obtained using a much smaller number of filters. 

One way to support the re-use of filters to represent multiple phase-shifts is 
by relaxing the diagonal factorization (Eq.~\ref{equation:diagonalfactorization}) with a factorization 
that allows for a richer connectivity:
\begin{equation} 
w_{ijk} = \sum_{def} C_{def} w^x_{id} w^y_{je} w^z_{kf}
\label{equation:fullfactorization}
\end{equation} 
where we $C_{def}$ are the components of a (non-diagonal) core tensor $C$. Note that,  
if $C$ is diagonal so that $C_{def}=1$ iff $d=e=f$, 
we would recover 
the PARAFAC factorization (Eq.~\ref{equation:diagonalfactorization}).
The energy function now turns into (cf. Eq. \ref{equation:energydiagonalfactorization}):
\begin{equation} 
E = \sum_{def} C_{def} \big( \sum_i w^x_{id} x_i \big) \big( \sum_j w^y_{je} y_j \big) \big( \sum_k w^h_{kf} h_k \big)
\label{equation:energyfullfactorization}
\end{equation} 
and inference into 
\begin{equation} 
p(h_k|\bm{x}, \bm{y}) = \sigma\big(\sum_{def} C_{def} w^h_{kf} \big( \sum_i w^x_{id} x_i \big) \big( \sum_j w^y_{je} y_j \big) \big)
\label{equation:energydiagonalfactorization3}
\end{equation} 

As the number of factors is typically large, a full matrix $C$ would be computationally too 
expensive. 
In fact, as we discuss in Section~\ref{section:fgbm} there is no other 
reason to project onto filters than reducing connectivity in the projected representation. 
Also, by the discussion in the previous section, there is very a strong inductive bias 
towards allowing \emph{groups of factors} to interact.  

This suggests using a core-tensor that allows features to come in groups of a fixed, small size, 
such that all pairs of filters within a group can provide products to mapping units. 
By the analysis in Section~\ref{section:fgbm}, training on translations 
or natural video is then likely to yield groups of roughly constant frequency and orientation and to 
differ with respect to phase.
We shall refer to this model as ``group-gating'' model in the following. 
As the values $C_{def}$ may be absorbed into the factor matrices and are learned from data, 
it is sufficient to distinguish only between non-zero and zero entries in $C$, and we 
set all non-zero entries to one in what follows. 

By defining the filter groups $\mathcal{G}_g, g=1,\ldots,G$, we can write inference in the model consequently as 
\begin{equation} 
p(h_k|\bm{x}, \bm{y})= 
\sigma\big(\sum_{g} \sum_{d\in \mathcal{G}_g} \sum_{e\in \mathcal{G}_g }w^h_{k, d \cdot |\mathcal{G}_g| + e} \big( \sum_i w^x_{id} x_i \big) \big( \sum_j w^y_{je} y_j \big) \big)
\label{equation:energydiagonalfactorization}
\end{equation} 
which is illustrated in Figure~\ref{figure:blocktensor} (right).
Note that each hidden unit can still pool over all pair-wise products between features. 
The overall number of feature products is equal to the number of groups times the 
group size. 
In practice, it makes sense to set the number of factors to be a multiple of the group size, 
so that all groups can have the same size.

A convenient way to implement the group-gating model is by computing all required products 
by creating $G$ copies of the input-factor matrix and $G$ copies of the output-factor matrix 
and permuting the columns (factors) of one of the two matrices appropriately. 
(Note in particular that when using a large number of factors, masking, i.e.\ forcing a subset of 
entries of $C$ to be zero, would not be feasible.)
It is possible to reduce the number of filters further by allowing for multiplicative interactions 
between only \emph{one} filter per group from the input image and all filters from the output 
image (or vice versa). 
This leads to an asymmetric model, where the number of filters is not equal for both images.

\subsection{Significance for square-pooling models}
It is interesting to note that the same analysis 
applies to the responses of energy-model complex cells (e.g., \cite{HyvarinenHI00, mcrbm}), too, 
if images and filters are contrast normalized \cite{multiview}. 
In this case, the response of the energy model is the same as a factored GBM mapping unit applied 
to a single image, i.e.\ $\bm{x}=\bm{y}$ (see, for example, \cite{mcrbm, multiview}). This shows 
that the gating interpretation of frequency/orientation groups and topographic structure 
applies to these models, too. 

In the same way, we can interpret a square-pooling model applied to a \emph{single image} as 
an encoding of the relationship between, say, the rows or the columns within the patch. 
In natural images, the predominant transformation that takes a row to the next row is 
a local translation. The emergence of oriented Gabor features can therefore also be 
viewed as the result of modeling these local translations.

\section{Experiments}
\label{section:experiments}
{\bf Cross-correlation models vs. energy models:} 
\label{section:mcevsgae}
\begin{figure}
\begin{center}
\begin{tabular}{cc}
    {\includegraphics[angle=0, width=5.3cm]{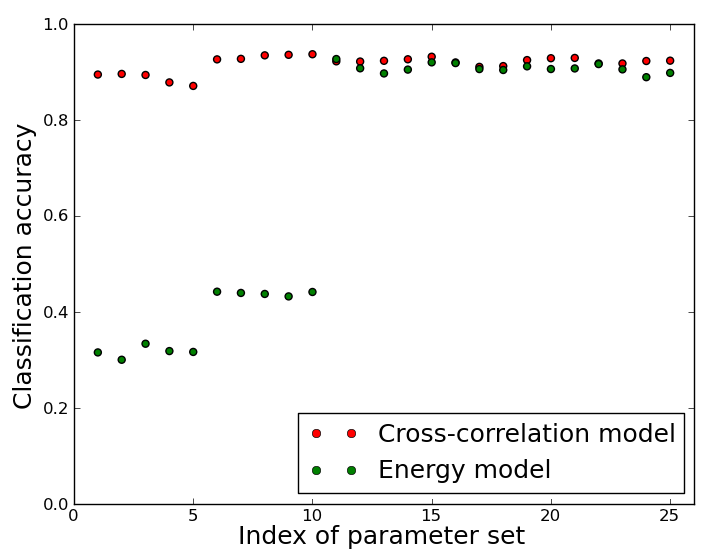}}&
    {\includegraphics[angle=0, width=5.3cm]{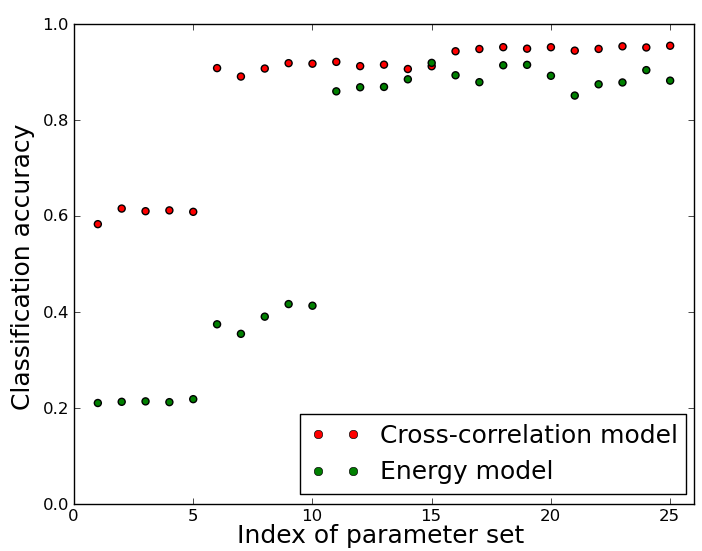}}
\end{tabular}
\end{center}
\caption{Classification accuracy of a Gated Boltzmann Machine vs. a cRBM on shifts {\bf (left)} and rotations {\bf (right)}. A total of 25 different sets of parameters were tested, with the numbers of filters and mapping units being varied between parameter sets.}
\label{figure:mce_vs_gae}
\end{figure}
To get some quantitative insights into the relationship between cross-correlation 
and energy models, we compared the standard Gated Boltzmann Machine with a 
square-pooling GBM \citep[e.g.][]{mcrbm,higherordergradientbased} trained on the
concatenation of the images.
We used the task of classifying transformations from the mapping units, using 
transformations for which we can control the ground truth.
The models are trained on image patches of size $13 \times 13$ pixels, 
which are cropped from larger images to ensure that no border artifacts are introduced. 
Training-, validation- and test-data contain $100.000$ cases each. 
We use logistic regression to classify transformations from mapping units, where 
we determine the optimal number of mapping units and factors, as well as the learning rates 
for the transformation models on the validation data. 
Using random images ensures that single images contain no information about the 
transformation; in other words, a single image cannot be used to predict the transformation.

We compare the models on translations and rotations. 
Shifts are drawn uniform-randomly in the range $[-3, 3]$ pixels. 
We used four different labels corresponding to the four quadrants of motion direction. 
Rotations are drawn from a \emph{von Mises} distribution (range $[-\pi, \pi]$ rad), which we 
scaled down to a maximal angle of 36$^\circ$.
We used $10$ labels by equally dividing the set of rotation angles. 

The results are shown in Figure~\ref{figure:mce_vs_gae} and they demonstrate that 
both types of model do a reasonably good job at prediction the transformations from 
the image pairs. 
The experiment verifies the approximate equivalence of the two types of model 
derived, for example, in \cite{FleetBinocular,multiview}. 
The cross-correlation model does show slightly better performance than 
the energy model for large data-set sizes, and the difference gets more pronounced as 
the training dataset size is decreased, 
which is also in line with the theory. 
Energy models have been the standard approach to feature grouping in 
the past \cite{NaturalImageStatistics}. 

{\bf Learning simple transformations:} 
\begin{figure}
\begin{center}
\begin{tabular}{ccc}
    {\includegraphics[angle=0, width=4.cm]{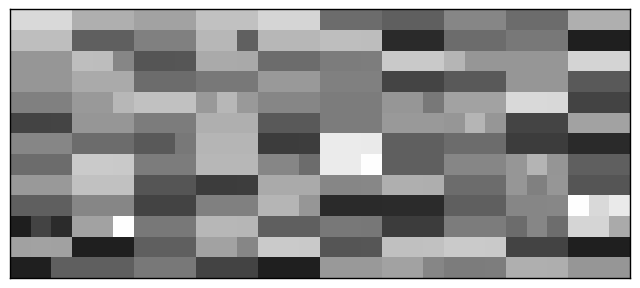}} &   
    {\includegraphics[angle=0, width=4.cm]{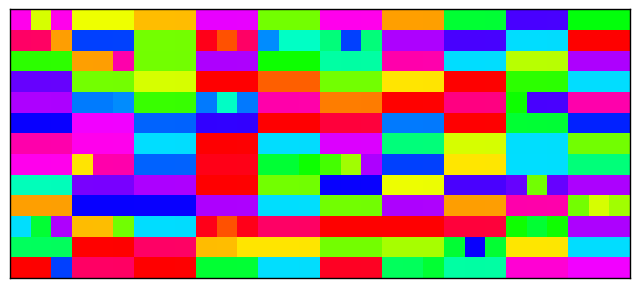}} &
    {\includegraphics[angle=0, width=4.cm]{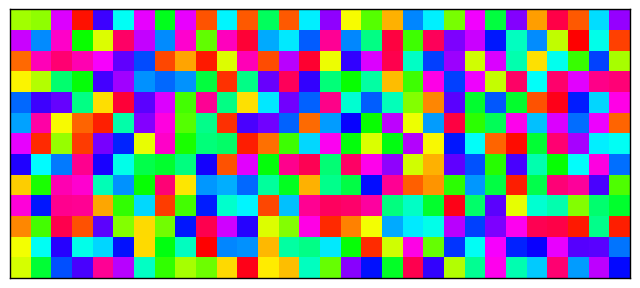}} \\
    {\includegraphics[angle=0, width=4.cm]{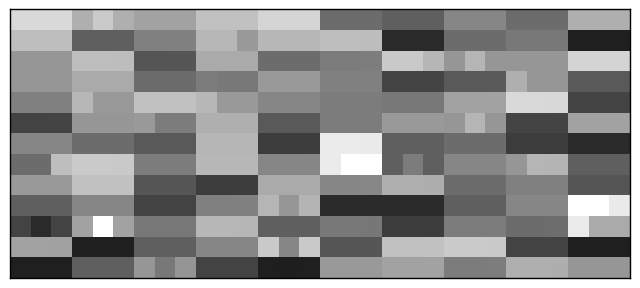}} &
    {\includegraphics[angle=0, width=4.cm]{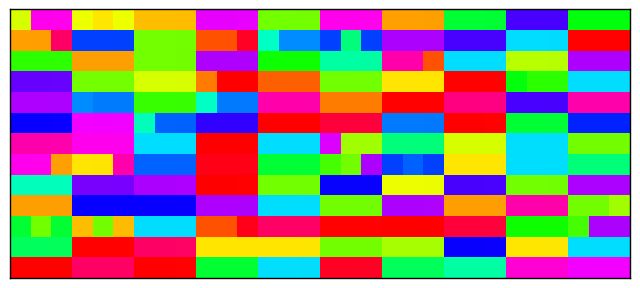}} &
    {\includegraphics[angle=0, width=4.cm]{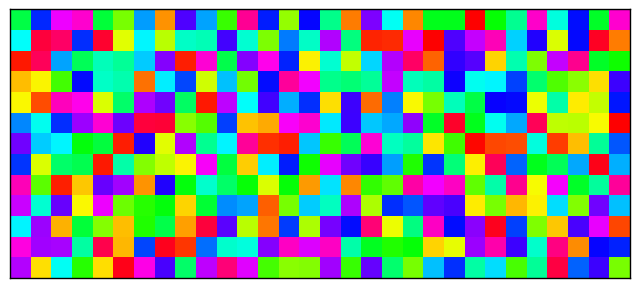}}
\end{tabular}
\end{center}
\caption{Frequency {\bf (left)}, orientation {\bf (middle)} and phase {\bf (right)} of filters trained on translated image pairs, with filters applied to the input images in the top and filters for the output image on the bottom. A group size of $3$ is used here, which means each three consecutive input filters and their corresponding output filters form a group. Notice that frequency and orientation are often identical for all filters within a group, whereas the phase varies. Best viewed in color.}
\label{figure:property_plots-group_size_3}
\end{figure}
We trained the group-gating model with group size $3$, $128$ mapping units and $392$ filters on 
translating random-dot images of size $13\times 13$ pixels. 
Figure \ref{figure:property_plots-group_size_3} shows three pairs of plots, where
the left, center and right pair depict the dominant frequencies, orientations and phases of
the filters, respectively. We extracted these from the filters using FFT. 
Each pair shows properties of input (top) and output filters (bottom).

\begin{figure}
\begin{center}
\begin{tabular}{ccc}
    {\includegraphics[angle=0, width=2.3cm]{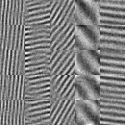}}&
    {\includegraphics[angle=0, width=2.3cm]{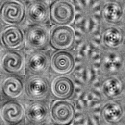}}&
    {\includegraphics[angle=0, width=7.0cm]{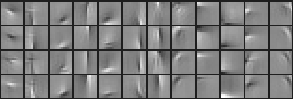}}
\end{tabular}
\end{center}
\caption{Filter-groups learned from translations {\bf (left)} and rotations {\bf (middle)} using a group-gating model with group size~$5$. Each group corresponds to one column in the plot. 
{\bf Right:} Filter-groups learned from natural videos, using a group-gating model with group size~$4$. Only output filters are shown. Frequency, orientation and position are constant, while phase varies, within each group.}
\label{figure:hateren_groupsof4}
\end{figure}

Within an image, each filter is represented by a single uni-colored square. For
frequency plots, the squares are in gray-scale with the brightness corresponding
to the frequency of the filter (black represents frequency zero, white is the highest
frequency across all filters); in the orientation and phase plots, the
squares differ in color, where the angle determines the color according to the
HSV representation of RGB color space. 
The figures confirm that filter-groups tend to be homogeneous with respect to 
frequency and orientation\footnote{In some groups, there appear to be some outliers, whose frequency 
or orientation does not match the other filters in the group. 
These are typically the result of spurious maxima in the FFT amplitude 
due to small irregularities in the filters.}.
Contrary to that, phases differ within each group, as expected.
The same can be seen in Figure~\ref{figure:hateren_groupsof4} (left and middle), which shows 
subsets of filter groups learned from translations and rotations of random images, 
using patchsize $25\times 25$ and group size $5$. 

{\bf Learning transformation features from natural videos:} 
\label{section:haterengroups}
Figure~\ref{figure:hateren_groupsof4} (right) shows a subset of filter groups learned from 
about one million patches of size $12 \times 12$ pixels that we crop from 
the \emph{van Hateren} broadcast television database \cite{hateren98independent}.
We trained the asymmetric model to predict each frame from its predecessor. 
All patches were PCA-whitened, retaining $95\%$ of the variance. We
trained a model with $128$ mapping units and $256$ filters, and we used a group size of $4$. 
The figure shows that the model learns Gabor features, where frequency, orientation and position 
are nearly constant within groups, and the set of possible phases is covered relatively 
evenly per group. 

{\bf Quantitative results for group-gating:}
Classification accuracies for the same model and data as described in Section~\ref{section:mcevsgae}
are reported in Table~\ref{table:mce_vs_gae}. 
The equivalent number of group-gating filters is shown in parentheses, where equivalence 
means that the group-gating model has the same number of parameters in total as the factored GBM. 
This makes it possible to obtain a comparison that is fair in terms of pure parameter count 
by comparing performance across the columns of the tables. 
However, the table also shows that even along columns, the group-gating model robustly 
outperforms the factored GBM, except when using a very small number of factors. In this case,  
the parameter-equivalent number of $44$ factors is probably too small to represent the transformations with
a sufficiently high resolution. 

\begin{table}[t]
\vskip 0.15in
\begin{center}
\begin{small}
\begin{sc}
\begin{tabular}{ccccc}
 &\multicolumn{2}{c}{Translation}&\multicolumn{2}{c}{Rotation}\\
 Number of filters & Gating & Group Gating &Gating &Group Gating \\
 81 (44)   & \textbf{89.69\%} & 81.68\%  & \textbf{90.08\%} & 89.06\% \\
 225 (121) & 92.40\% & \textbf{92.59\%}  & 91.50\% & \textbf{94.65\%} \\
 441 (237)  & 92.20\% & \textbf{93.36\%} &   90.76\% & \textbf{95.65\%}\\
 729 (392) & 91.44\% & \textbf{93.46\%}  & 91.71\% & \textbf{95.51\%}\\
 900 (483) & 91.29\% & \textbf{93.20\%}  & 89.95\% & \textbf{95.00\%} \\
\end{tabular}
\end{sc}
\end{small}
\end{center}
\vskip -0.1in
\caption{Classification accuracy of the standard and group-gating models on image pairs showing translations and rotations.}
\label{table:mce_vs_gae}
\end{table}

\begin{figure*}
\begin{center}
\begin{tabular}{cc}
    \hspace{-0.3cm}
    {\includegraphics[angle=0, width=6.9cm]{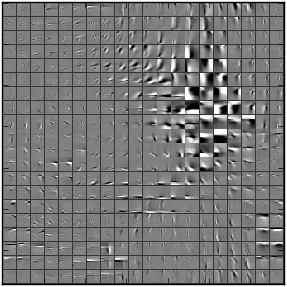}}&
    \hspace{-0.3cm}
    {\includegraphics[angle=0, width=6.9cm]{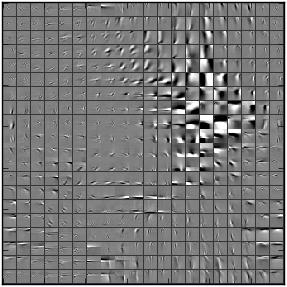}}
\end{tabular}
\end{center}
\caption{Topographic filter maps learned by laying out features in two dimensions 
and allowing for multiplicative interactions among nearby units.
{\bf Left:} Input filters, {\bf Right:} Output filters.}
\label{figure:hateren_topo}
\end{figure*}

\subsection{Topographic feature maps from local gating}
We also experimented with overlapping group structures by 
letting $\mathcal{G}_g$ share filters. This makes it necessary to 
define which filters are shared among which groups. 
A convenient approach is to lay out all features (``simple cells'') 
in a low-dimensional space and to define groups over all those units which 
reside within some pre-defined neighborhoods, such as in a grid of 
size $n \times n$ units (e.g., \cite{NaturalImageStatistics}). 

Figure~\ref{figure:hateren_topo} shows the features learned from 
the \emph{van Hateren} data (cf., Section~\ref{section:haterengroups}) with 
patchsize $16\times 16$ and $99.9\%$ variance retained after whitening. 
We used $400$ filters that we arranged in a $2$-D grid (with wrap-around) 
using a group size of $5\times 5$. 
We found that learning is simplified when we also initialize the factor-to-mapping 
parameters so that each mapping unit has access to the filters of only one group.

The figure shows how learned filters are arranged in two topographic feature maps 
with slowly varying frequency and orientation within each map. 
It also shows how low-frequency filters have the tendency of being 
grouped together \cite{NaturalImageStatistics}.
From the view of phase analysis (see Section~\ref{section:fgbm}), 
the topographic organization is a natural result of imposing an additional constraint: 
Filter sets that come in similar frequency, orientation and position  
are now forced to share coefficients with those of neighboring groups. 
The simplest way to satisfy this constraint is by having nearby groups be 
\emph{similar} with respect to the properties they share. 
The apparent randomness of phase simply follows from the requirement that 
multiple phases be present within each neighborhood. 
It is interesting to note that using shared group structure is 
equivalent to letting units that are nearby in space affect each other 
multiplicatively. 
Localized gating may provide a somewhat more plausible explanation 
for the emergence of pinwheel-structures than squaring non-linearities, 
which are used, for example, in subspace models 
or topographic ICA (see \cite{HyvarinenISA, HyvarinenHI00}).

\section{Conclusions}
\label{section:conclusions}
Energy mechanisms and ``square-pooling'' are common approaches to modeling feature 
dependencies in sparse coding, and to learn group-structured or invariant dictionaries.  
In this work we revisited group-structured sparse coding from the perspective of 
learning image motion and disparity from local multiplicative interactions. 
Our work shows that the commonly observed constancy of frequency and orientation of 
filter-groups in energy models 
can be explained as a result of representing transformations with local, multiplicative feature gating. 
Furthermore, topographically structured representations (``pinwheels'') can emerge 
naturally as the result of binocular or spatio-temporal learning that utilizes 
spatially constrained multiplicative interactions. 
Our work may provide some support for the claim that localized multiplicative interactions 
are a biologically plausible alternative to square pooling for implementing 
stereopsis and motion analysis \cite{MelSingleCell}. 
It may also help explain why the development of pinwheels in V1 may be tied 
the presence of binocular vision and why topographic organization of features 
does not appear to occur, for example, in rodents \cite{stevens2011universal}.  

\begin{small}

\bibliography{xgae}
\bibliographystyle{plain}

\end{small}

\end{document}